\documentclass[sigplan,screen,nonacm,review=false,timestamp=false]{acmart}
\usepackage{graphicx}
\usepackage[utf8]{inputenc}
\usepackage[T1]{fontenc}
\usepackage{mathptmx} 
\usepackage{caption}
\usepackage{subcaption}
\graphicspath{ {./images/} }

\AtBeginDocument{%
  \providecommand\BibTeX{{%
    \normalfont B\kern-0.5em{\scshape i\kern-0.25em b}\kern-0.8em\TeX}}}

\acmConference[ICIMMI '22]{International Conference on Information Management & Machine Intelligence}{December 23--24, 2022}{Jaipur, India}

\begin{document}

\title{Towards Edge-Cloud Architectures for Personal Protective Equipment Detection}

\author{Jarosław Legierski}
\email{jaroslaw.legierski@orange.com}
\affiliation{%
  \institution{Orange Innovation, Orange Polska~S.A.}
  \city{Warsaw}
  \country{Poland}
  \postcode{00-001}
}

\author{Kajetan Rachwał}
\email{kajetan.rachwal@ibspan.waw.pl}
\author{Piotr Sowinski}
\email{piotr.sowinski@ibspan.waw.pl}
\affiliation{%
  \institution{Systems Research Institute Polish Academy of Sciences}
  \country{Poland}
}
\affiliation{
  \institution{Warsaw Univesity of Technology}
  \city{Warsaw}
  \country{Poland}
}

\author{Wojciech Niewolski}
\email{wojciech.niewolski@orange.com}
\affiliation{%
  \institution{Orange Innovation, Orange Polska~S.A.}
  \country{Poland}
 }
\affiliation{
  \institution{Warsaw Univesity of Technology}
  \city{Warsaw}
  \country{Poland}
}

\author{Przemysław Ratuszek}
\email{przemyslaw.ratuszek@orange.com}
\author{Zbigniew Kopertowski}
\email{zbigniew.kopertowski@orange.com}
\affiliation{%
  \institution{Orange Innovation, Orange Polska~S.A.}
  \city{Warsaw}
  \country{Poland}
 }

\author{Marcin Paprzycki}
\email{marcin.paprzycki@ibspan.waw.pl}
\author{Maria Ganzha}
\email{maria.ganzha@ibspan.waw.pl}
\affiliation{%
  \institution{Systems Research Institute Polish Academy of Sciences}
  \city{Warsaw}
  \country{Poland}
}

\renewcommand{\shortauthors}{Legierski and Rachwał et al.}

\begin{abstract}
Detecting Personal Protective Equipment in images and video streams is a relevant problem in ensuring the safety of construction workers. 
In this contribution, an architecture enabling live image recognition of such equipment is proposed. The solution is deployable in two settings -- edge-cloud and edge-only.
The system was tested on an active construction site, as a part of a larger scenario, within the scope of the ASSIST-IoT H2020 project.
To determine the feasibility of the edge-only variant, a model for counting people wearing safety helmets was developed using the YOLOX method.
It was found that an edge-only deployment is possible for this use case, given the hardware infrastructure available on site.
In the preliminary evaluation, several important observations were made, that are crucial to the further development and deployment of the system.
Future work will include an in-depth investigation of performance aspects of the two architecture variants.
\end{abstract}


\maketitle

\section{\label{sec:intro}Introduction}
Nowadays, the demand for intelligent video analytics is growing across a wide spectrum of application areas~\cite{zhang2019edge}.
The key part of such systems is usually an image recognition (IR) component. However, as of today, the IR subsystem is, most commonly, deployed in the cloud.
This approach offers multiple benefits, such as availability of large and scalable computational resources, reliable APIs, and shifting the burden of system maintenance to the cloud service provider.
However, this comes at a cost. Sending data to the cloud raises both security and privacy concerns.
Moreover, communicating with the cloud always induces network latency, which may be significant in time-critical applications.
To address issues brought about by cloud-centric solutions, edge computing has been proposed. Here, the core of the approach is processing the data as close to the source as possible.
This allows for latency reduction, and helps ensure the security and privacy of data, which remains within the local network.
However, edge computing has its own set of issues. Typically, the computational resources, which are available at the edge are considerably smaller.
A possible solution to addressing the downsides of both these options is a combined approach -- an \emph{edge-cloud continuum}, where data is partially processed on the edge and partially in the cloud.
However, this raises the obvious question: at which point(s), within the continuum, individual parts of the system should be deployed.

Here, this question is considered within a real-world scenario of monitoring the entrance to an active construction site.
Specifically, the system is tasked with ensuring that (1) no unauthorized people enter the worksite, and (2) everybody is wearing appropriate Personal Protective Equipment (PPE), i.e. helmets and safety vests.
The scenario is evaluated as part of the ASSIST-IoT project, on a construction site in Warsaw, Poland, managed by the construction company Mostostal Warszawa. Here, the edge versus cloud discussion becomes particularly relevant. On the one hand, the privacy of workers is of paramount importance, while latencies must be minimized, to ensure a quick reaction, which hints at an edge deployment. On the other hand, given the limited hardware resources available on the edge, and the extremely harsh conditions of the construction site, a cloud deployment seems attractive.

Given the possible benefits of both solutions, in this contribution, a solution is proposed for an edge-cloud continuum video analytics architecture. The architecture can be deployed in two variants (edge-only, and edge-cloud), described in the \emph{Architecture} section.
Moreover, to determine the viability of the solution, an initial experimental study was performed. Here, an IR model was developed and integrated with the edge-only variant of the architecture. Next, it was tasked with detecting when personnel wearing PPE entered and exited the work site.

\section{\label{sec:background}Background}

To provide a context for this study, the state of the art of (1) IR system architectures and (2) machine learning models for PPE detection is summarized.

\paragraph{System architectures}

The most obvious benefit of deploying IR systems on the edge is the decreased latency.
This was demonstrated in~\cite{Zhou2020}, where facial recognition models were deployed on the edge.
The authors found that deploying the models on the edge resulted in significantly better response speeds, as compared to a cloud deployment.
In other studies~\cite{Liu2018, Liu2019}, the viability of deploying deep convolutional neural networks (CNNs) in the edge-only scenario was investigated.
CNNs are characterized by high resource utilization, and thus are typically deployed in the cloud.
The studies found that deploying CNNs is viable on mobile devices, when parts of the computation can be offloaded to other edge devices.
Edge deployment allowed to achieve a consistently low latency of 2.24\,ms while using CNNs to perform real-time object tracking in augmented reality~\cite{Liu2019}.
Both studies showed that the edge deployment distributing the workload increased the inference capabilities of the system, as the models could not be run on the disconnected mobile devices alone.

One study~\cite{Huang2017} investigated an edge-cloud architecture, where data preprocessing servers were deployed close to the data source. The preprocessed data was then sent to a cloud-based deep-learning platform.
This resulted in decreasing network latency and traffic.
It also increased the security and privacy of the raw data.

On the other hand, edge deployments are more limited in terms of the available hardware. Low computational resources naturally limit the size of models and inference speed.
A study compared different implementations (based on TensorFlow, TensorRT, and TFLite) of the same video processing model~\cite{Koubaa2022}, and found them to differ in their resource utilization. The choice of implementation influenced the energy consumption of the model, as well as its inference speed.
Interestingly, the slowest implementation (TFLite) was the most energy efficient.
It was also found that TFLite managed to remain on par with the other implementations in terms of speed, when processing low-resolution video.
In the case of high-resolution video, more resource-intensive models were needed to maintain the speed, suggesting that a cloud deployment could be more beneficial in low-resource settings.
Nevertheless, some resource-intensive models can be deployed on the edge, if resources available there are sufficient.
The deployment proposed in a different study required all nodes to be equipped with a GPU~\cite{Liu2018}.
This allowed the authors to use CNNs on the edge.
A similar result was reported in~\cite{JL-2}, were IR models deployed on a Raspberry Pi 4B, equipped with a camera, and an Intel Neural Compute Stick 2 (a USB device for deep learning inference on the edge) were studied.
These devices were chosen for their low power consumption and good computing capabilities.
Overall, a model tasked with detecting PPE in the form of helmets and safety vests achieved precision on the order of $99.5\%$.

\paragraph{Models for PPE detection}

Effective video analytics-based methods for detecting the presence of protective helmets, worn by workers, due to its health and safety importance, is currently a hot research topic. 

The usage of existing, unmodified machine learning models for detecting protective head covers does not provide sufficient detection accuracy, as proven in a recent study~\cite{WN-1-E-12}. In said article, several versions of the popular YOLO algorithm~\cite{bochkovskiy2020yolov4} were compared. It was shown that the most effective version of YOLO for helmet detection is the v4. After improving the loss function, it achieved more than 93\% accuracy during tests. A similar study~\cite{WN-2-E-13} focused on improving the YOLOv5 algorithm. The system achieved results close to 97\% accuracy, thanks to the improvement of the structure of the neural network. Another study~\cite{WN-3-E-14}, also investigated improving YOLOv5. However, instead of the algorithm itself, work was focused on processing of input data by applying filters on the input image. This allowed to improve the accuracy to above 95\%.
Yet another study~\cite{ZK-2-17} presented an approach for improving the detection speed and accuracy by designing a multi-level pyramidal feature fusion network based on the ConCaNet attention mechanism. Here, YOLOv3 was applied and a dataset with 6000 images was used. The results demonstrate the effectiveness of this approach, which managed to reduce the number of necessary parameters.

Helmet detection can also be done using the SSD-MobileNet algorithm~\cite{Howard2017}, which is based on yet another variant of CNN. An analysis of this method, reported in~\cite{WN-4-E-15}, tested its effectiveness and managed to reach 80\% accuracy during tests. In a wider comparison of algorithm types~\cite{WN-5-E-11}, the authors proposed a helmet detection method based on a dynamically changing neural network -- SHDDM (Safety Helmet Detection Dynamic Model). The developed model analyzes the human posture and defines the area where the helmet should be located, to eliminate the detection of the helmet outside the head area and thus reduce the false positive rate. There are also other approaches to helmet detection, such as methods based on color and shape used to to locate the face, and the proper wearing of a helmet~\cite{ZK-1-16}. Another solution used low-resolution images, captured from a video stream, using the Local Binary Pattern (LBP) and gray-level co-occurrence matrix (GLCM) methods along with a back-propagation neural network~\cite{ZK-4-19}.

Another study~\cite{PR-4} investigated the usefulness of artificially created images in the training of CNNs for PPE detection. The paper presented the results achieved with YOLOv3, trained on artificial images generated by the Rockstar Advanced Game Engine (RAGE) from the Grand Theft Auto V video game. This approach achieved a mean average precision (mAP) of only 55.11\% on a test dataset consisting of real-world images. The mAP for synthetic images was much higher at 87.24\%. It should be noted that the poor results for the real-world images are most likely caused by the RAGE engine being unable to generate a sufficient amount of head, welding mask, ear protection, and chest object variations.

As can be seen, there are many possibilities for detecting protective helmets. Here, the SHDDM is particularly noteworthy, as it has an important feature of checking whether the helmet is worn properly, and not only detecting its presence. This, in turn, is particularly relevant in real-world applications.

\section{\label{sec:architecture} Proposed Architecture}

\begin{figure}[htp]
    \centering
    \begin{minipage}{.49\textwidth}
        \includegraphics[width=.98\linewidth]{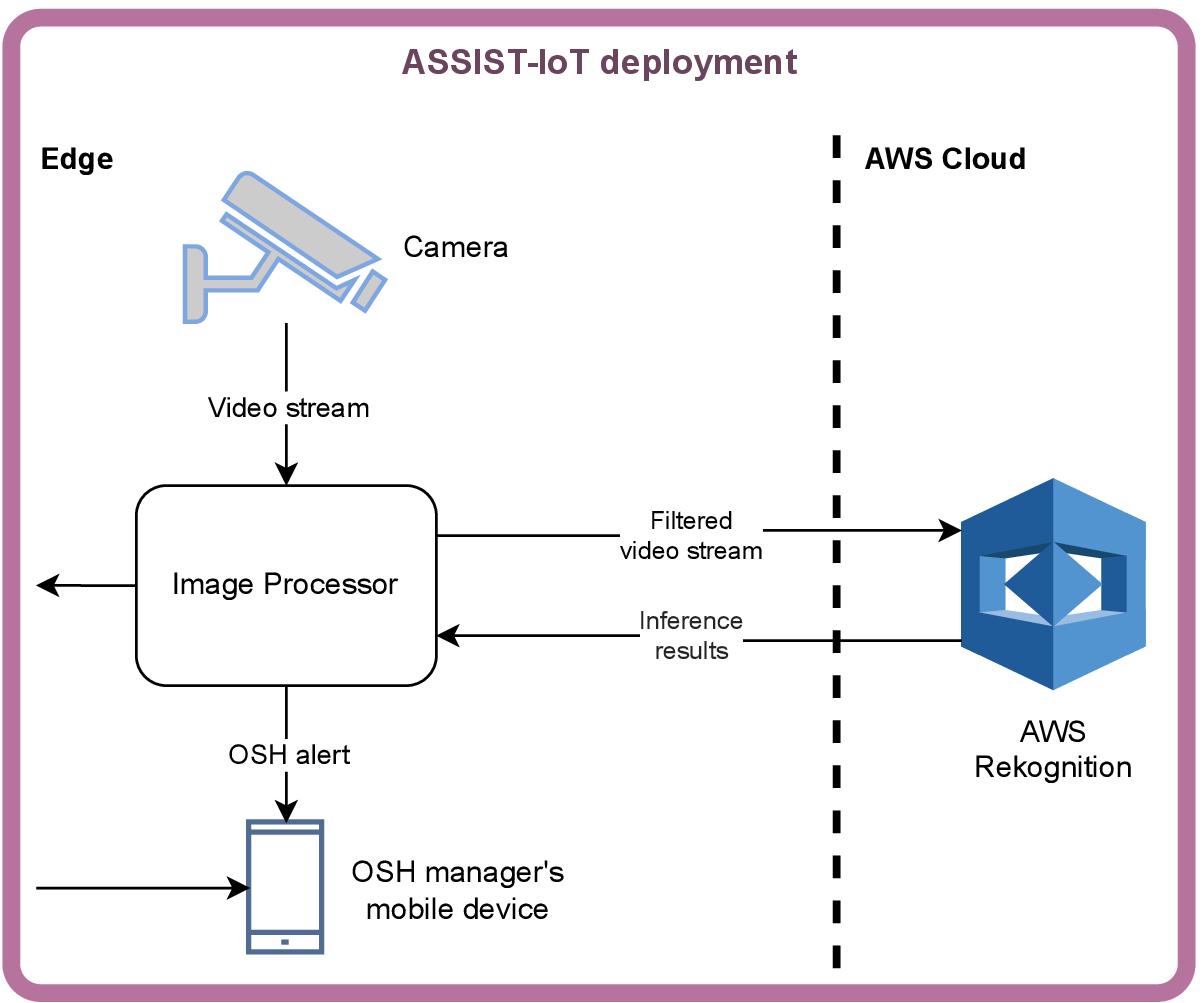}
        \captionof{figure}{Edge-cloud deployment}
        \label{fig:acloud}
    \end{minipage}
    \begin{minipage}{.49\textwidth} 
        \includegraphics[width=.98\linewidth]{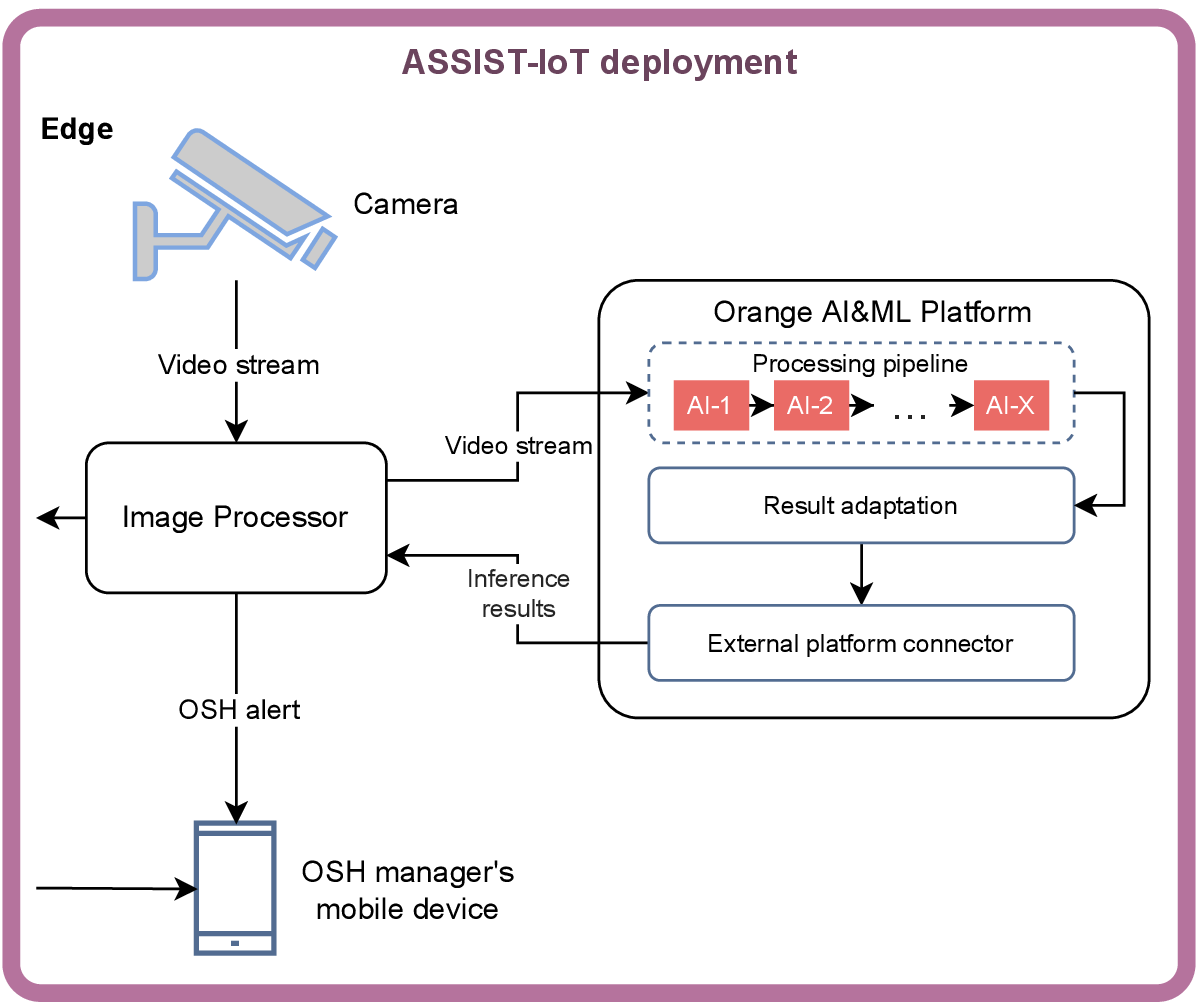}
        \captionof{figure}{Edge deployment}
        \label{fig:aedge}
    \end{minipage}
\end{figure}

The proposed video analytics system can be deployed in two architecture variants: edge-cloud (Fig.~\ref{fig:acloud}) and edge-only (Fig.~\ref{fig:aedge}). As outlined above, there are reasons to believe that both variants may be appropriate for the considered scenario.
Both architectures share a common core deployed on the edge, consisting of: a camera, the Image Processor (IP) component, and the OSH (Occupational Safety and Health) manager's mobile device.

The camera (in the reported experiments the Dahua IPC-HFW5449T-ASE-LED was used) provides a live RTSP video stream, which is directed to the Image Processor. 
The IP is a service written in Python, which can optionally perform preliminary image analysis.
Using configurable methods such as motion detection and brightness thresholding, the IP is able to discard image frames that do not contain moving people, reducing network traffic to components involved in actual image analysis.
It is also responsible for communicating with the rest of the system, designed in accordance with the ASSIST-IoT reference architecture~\cite{fornes2022assist}.
IP communicates with the rest of the system publishing alerts to an MQTT topic.
This design allows other components and devices in the ASSIST-IoT deployment to be notified in a streaming manner of any OSH violations, such as workers not wearing protective helmets.

In the first version of the architecture -- the edge-cloud deployment -- the IP is configured to use the cloud-based AWS Rekognition platform, with its PPE detection service.

In the edge-only variant, the video analysis is performed by the Orange AI\&ML Platform, 
which is deployed on a server on the construction site.
This edge deployment allows for maintaining lower network latency, and ensures the privacy of worker data. The
AI\&ML Platform's services are written as Python runnable modules that provide their own APIs and GUIs.
The services can reuse the APIs and GUIs provided by the platform, or build them from scratch.
A service collects frames from a video source, processes them in an ML pipeline specific to the service, and adapts or interprets the results.
The inference results from the Platform are forwarded to external services, with the use of provided connectors.
As the Orange AI\&ML Platform operates on the edge, all video processing takes place on the client's site, ensuring full security of customer data (video) and compliance with appropriate regulations, such as GDPR.

\section{\label{sec:method}Methodology}

\begin{figure}[htp]
    \centering
    \begin{minipage}{.49\textwidth}
        \centering
        \includegraphics[width=.8\linewidth]{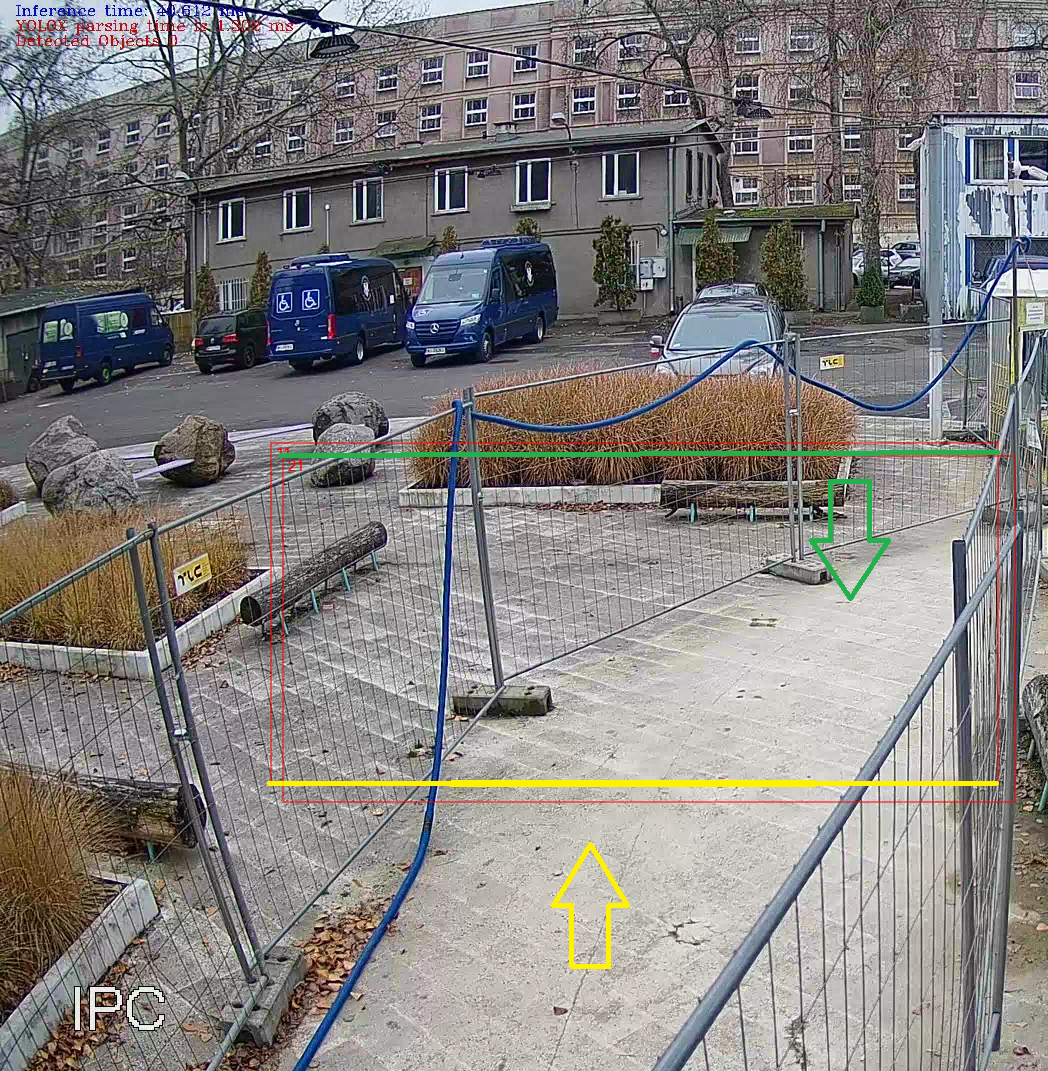}
        \captionof{figure}{Bounding boxes location on November 22 (before modification)}
        \label{fig:bbox_old}
    \end{minipage}
    \begin{minipage}{.49\textwidth} 
        \centering
        \includegraphics[width=.8\linewidth]{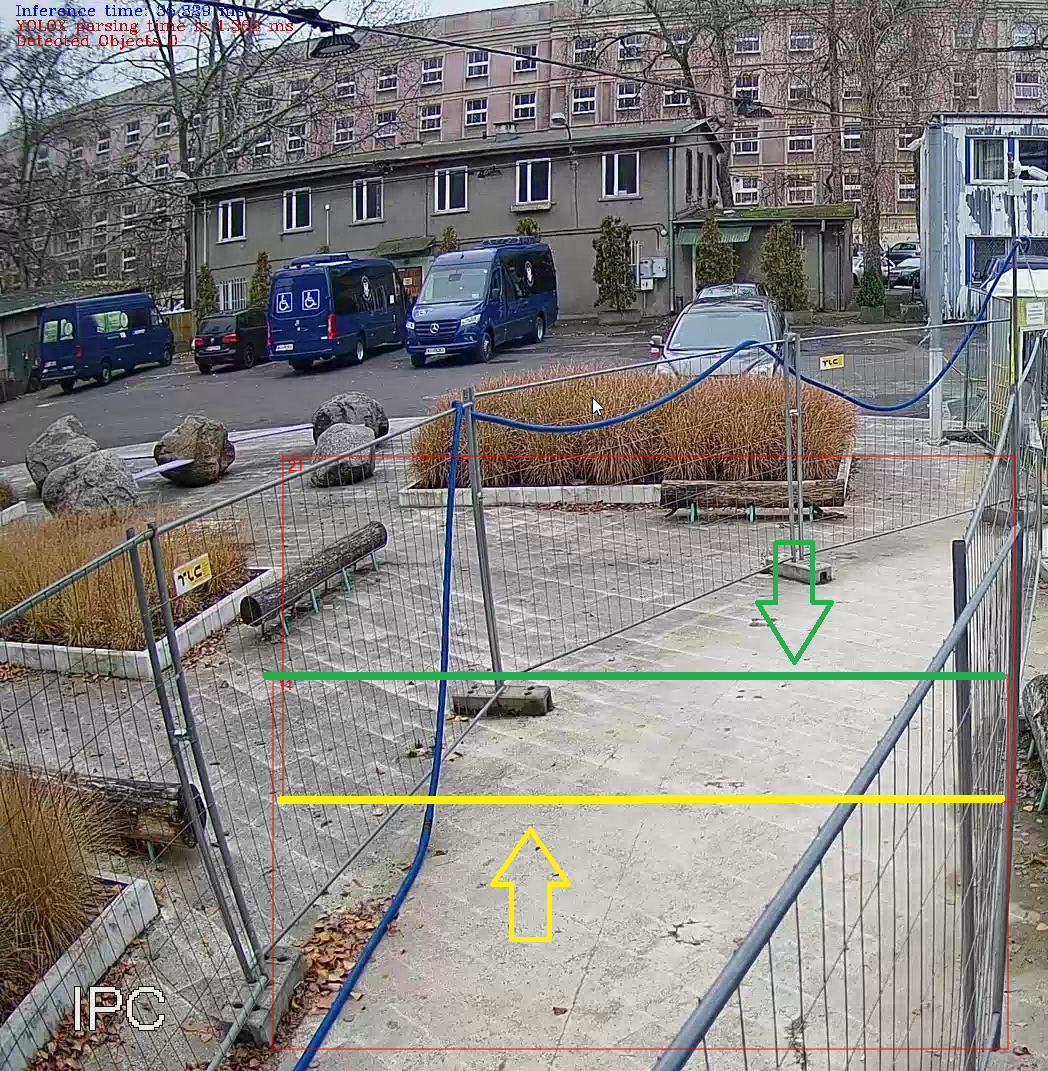}
        \captionof{figure}{Bounding boxes location on November 24 (after modification)}
        \label{fig:bbox_new}
    \end{minipage}
\end{figure}

As part of this study, a preliminary version of the edge-only variant of the architecture was deployed on an active construction site. Using the Orange AI\&ML Platform, a model was trained to count people wearing helmets entering and exiting a specific area.
The system counts people in helmets in defined recognition areas (bounding boxes), crossing the yellow and green lines visible in Figs.~\ref{fig:bbox_old} and~\ref{fig:bbox_new}. People entering the construction site are counted after crossing the green line, while people leaving are counted after crossing the yellow line. The machine learning pipeline consists of a YOLOX object detection model, trained for detecting heads in helmets, and a DeepSORT~\cite{Wojke2017} multi-object tracking algorithm.
The YOLOX model was trained using a dataset provided by the Northeastern University of China (\url{https://public.roboflow.com/object-detection/hard-hat-workers}). 

The system's results were compared to those obtained from an algorithm built into the Dahua camera. It should be noted that the camera counted \emph{all} people entering and leaving, including those without protective helmets. However, this should not impact the results much, as the safety regulations on this particular site forbid entering it without a helmet and the rule is strictly enforced before workers reach the counting location.

The measurements were performed in two series -- each using a different bounding box definition. A single series spanned the length of one workday on the construction site. The number of entering and leaving people was counted in hourly intervals (between 5 AM and 7 PM).

\section{\label{sec:results}Results}

The Tables~\ref{tab:bbox_old} and~\ref{tab:bbox_new} present the results of the performed experiments.
The Table~\ref{tab:bbox_old} contains measurements made on 22nd November 2022, with the bounding box set as presented in Fig.~\ref{fig:bbox_old}.
The average difference between the number of people entering, as measured by the camera and the model was equal to $-6.21$, with the standard deviation of $\sigma = 5.08$, whereas for people exiting it was $1.35$ and $\sigma = 2.73$ respectively. The correlation between entrances detected by the camera and the model deployed on the AI\&ML platform, expressed by the Pearson coefficient is $0.988$, whereas for exits $0.995$. The correlations were found to be statistically significant ($p \leq 0.05$).
Table~\ref{tab:bbox_new} contains measurements from 24th November 2022 (for modified detection areas, depicted in Fig.~\ref{fig:bbox_new}).
On that day, the average difference for entering was $-4.93$ with $\sigma = 4.25$ and for exiting $3.92$ with $\sigma = 4.92$. For these measurements the Pearson coefficient for people entering is equal to $0.993$ and exiting $0.989$. The correlations were found to be statistically significant ($p \leq 0.05$).

The tables also present differences in the number of people detected by the camera and the AI\&ML platform and the sum of these differences calculated for both movement directions: entries and exits.

\begin{figure}[]
    \centering
    \begin{minipage}{.49\textwidth}
        \centering
        \includegraphics[width=.8\linewidth]{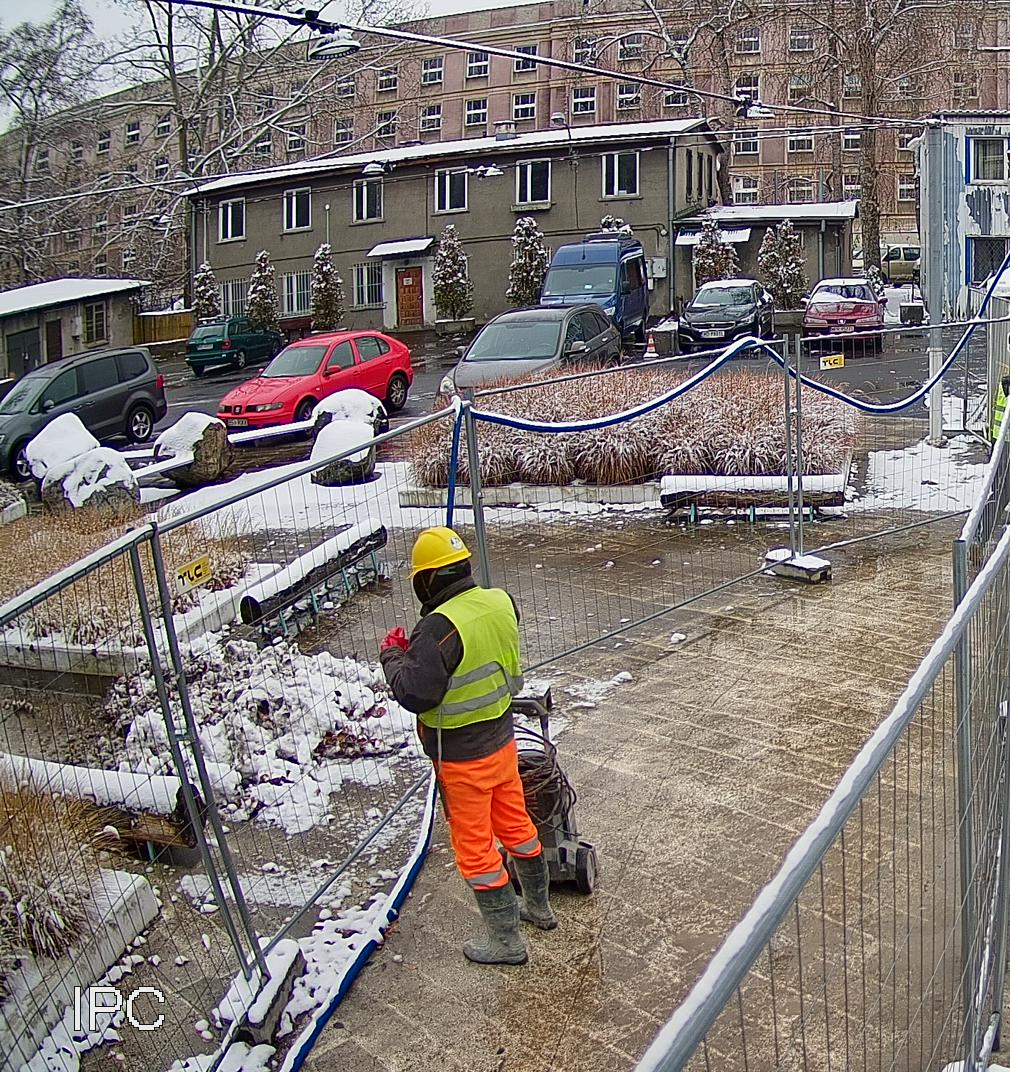}
        \captionof{figure}{Unexpected worker behavior -- staying in the detection area for longer}
        \label{fig:worker1}
    \end{minipage}
    \begin{minipage}{.47\textwidth} 
        \centering
        \includegraphics[width=.8\linewidth]{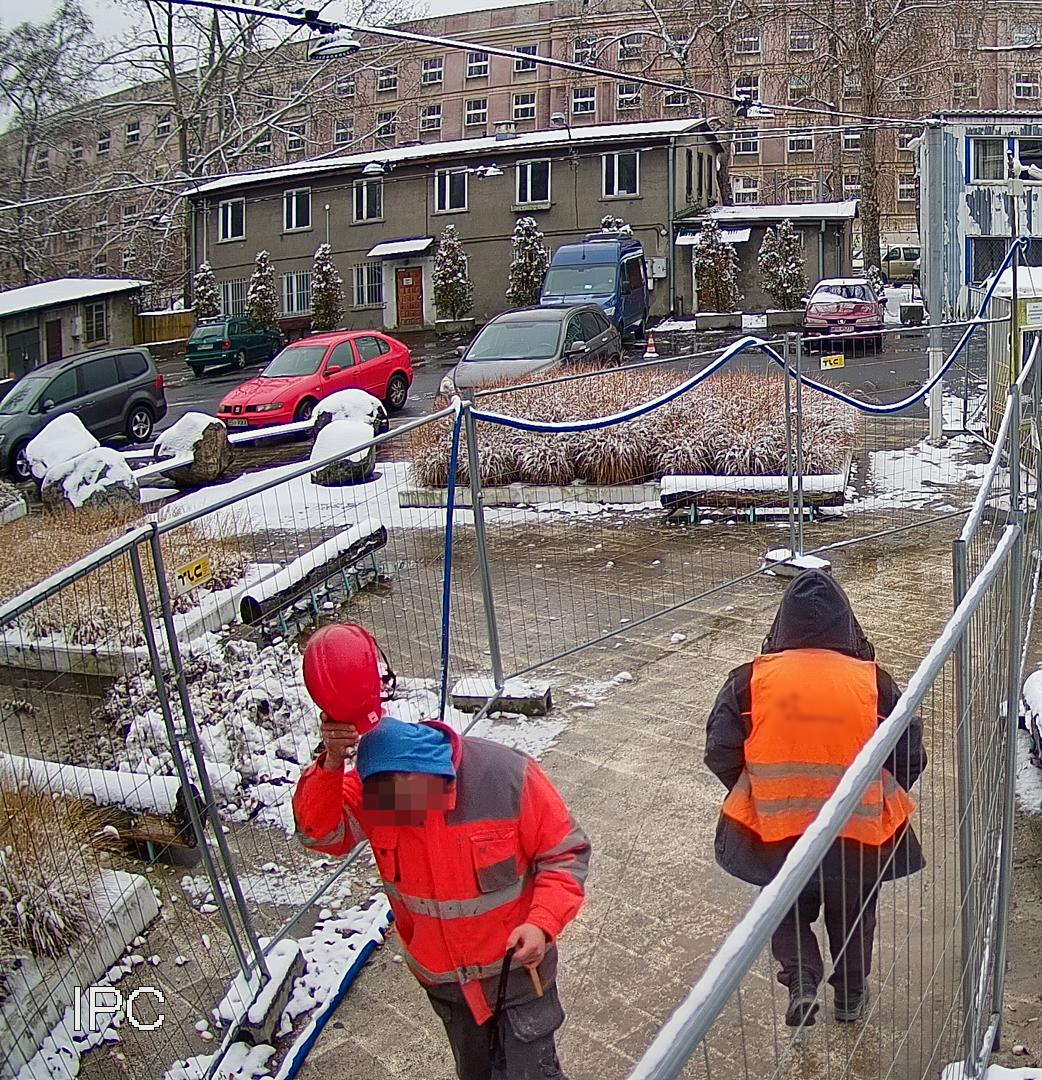}
        \captionof{figure}{Unexpected worker behavior -- putting the helmet on behind the detection line}
        \label{fig:worker7}
    \end{minipage}
\end{figure}

During the experiments, several unexpected events took place, which had a significant impact on the reported results.
Workers were observed acting in an unexpected manner -- lingering or walking around the detection area (Fig.~\ref{fig:worker1}).
It was also noticed that sometimes the workers put on their helmets after having passed the detection area (Fig.~\ref{fig:worker7}).
These behaviors present a challenge to the future system, as they significantly affect its accuracy.

\begin{table*}[t]
\resizebox{\textwidth}{!}{%
\begin{tabular}{|l|r|r|r|r|r|r|r|r|r|r|r|r|r|r|r|}
\hline
\textbf{Hour}      & \multicolumn{1}{c|}{05:00} & \multicolumn{1}{c|}{06:00} & \multicolumn{1}{c|}{07:00} & \multicolumn{1}{c|}{08:00} & \multicolumn{1}{c|}{09:00} & \multicolumn{1}{c|}{10:00} & \multicolumn{1}{c|}{11:00} & \multicolumn{1}{c|}{12:00} & \multicolumn{1}{c|}{13:00} & \multicolumn{1}{c|}{14:00} & \multicolumn{1}{c|}{15:00} & \multicolumn{1}{c|}{16:00} & \multicolumn{1}{c|}{17:00} & \multicolumn{1}{c|}{18:00} & \multicolumn{1}{l|}{\textbf{Total}} \\ \hline
\textbf{Dahua In}  & 12                     & 65                     & 84                     & 47                     & 26                     & 84                      & 50                      & 51                      & 28                      & 70                      & 28                      & 8                       & 9                       & 0                       & 562                        \\ 
\textbf{Dahua Out} & 2                      & 15                     & 21                     & 35                     & 81                     & 44                      & 61                      & 31                      & 63                      & 32                      & 59                      & 66                      & 26                      & 8                       & 544                        \\ \hline
\textbf{AI\&ML In}  & 11                     & 78                     & 87                     & 52                     & 28                     & 96                      & 60                      & 58                      & 43                      & 75                      & 33                      & 18                      & 10                      & 0                       & 649                        \\ 
\textbf{AI\&ML Out} & 2                      & 13                     & 23                     & 33                     & 73                     & 44                      & 63                      & 31                      & 58                      & 31                      & 59                      & 62                      & 25                      & 8                       & 525                        \\ \hline
\textbf{Diff. In}   & 1                      & -13                    & -3                     & -5                     & -2                     & -12                     & -10                     & -7                      & -15                     & -5                      & -5                      & -10                     & -1                      & 0                       & -87 \\ 
\textbf{Diff. Out}  & 0                      & 2                      & -2                     & 2                      & 8                      & 0                       & -2                      & 0                       & 5                       & 1                       & 0                       & 4                       & 1                       & 0                       & 19   \\ \hline
\end{tabular}}
\caption{Entries and exits to the construction site, 22 November 2022.}
\label{tab:bbox_old}
\end{table*}

\begin{table*}[t]
\resizebox{\textwidth}{!}{%
\begin{tabular}{|l|r|r|r|r|r|r|r|r|r|r|r|r|r|r|r|}
\hline
\textbf{Hour}      & {05:00} & \multicolumn{1}{c|}{06:00} & \multicolumn{1}{c|}{07:00} & \multicolumn{1}{c|}{08:00} & \multicolumn{1}{c|}{09:00} & \multicolumn{1}{c|}{10:00} & \multicolumn{1}{c|}{11:00} & \multicolumn{1}{c|}{12:00} & \multicolumn{1}{c|}{13:00} & \multicolumn{1}{c|}{14:00} & \multicolumn{1}{c|}{15:00} & \multicolumn{1}{c|}{16:00} & \multicolumn{1}{c|}{17:00} & \multicolumn{1}{c|}{18:00} & \multicolumn{1}{l|}{\textbf{Total}} \\ \hline
\textbf{Dahua In}   & 4                          & 57                         & 113                        & 62                         & 34                         & 73                         & 75                         & 65                         & 56                         & 93                         & 27                         & 10                         & 9                          & 0                          & 678                                 \\ 
\textbf{Dahua Out}  & 0                          & 10                         & 29                         & 57                         & 80                         & 53                         & 74                         & 41                         & 82                         & 52                         & 49                         & 84                         & 20                         & 8                          & 639                                 \\ \hline
\textbf{AI\&ML In}  & 3                          & 61                         & 113                        & 68                         & 43                         & 76                         & 79                         & 73                         & 69                         & 98                         & 33                         & 21                         & 10                         & 0                          & 747                                 \\ 
\textbf{AI\&ML Out} & 0                          & 11                         & 26                         & 46                         & 64                         & 48                         & 72                         & 38                         & 73                         & 49                         & 47                         & 82                         & 22                         & 6                          & 584                                 \\ \hline
\textbf{Diff. In}   & 1                          & -4                         & 0                          & -6                         & -9                         & -3                         & -4                         & -8                         & -13                        & -5                         & -6                         & -11                        & -1                         & 0                          & -69                                 \\ \
\textbf{Diff. Out}  & 0                          & -1                         & 3                          & 11                         & 16                         & 5                          & 2                          & 3                          & 9                          & 3                          & 2                          & 2                          & -2                         & 2                          & 55                                  \\ \hline
\end{tabular}}
\caption{Entries and exits to the construction site, 24 November 2022.}
\label{tab:bbox_new}
\end{table*}

\section{\label{sec:conclusions}Concluding remarks}

The tested model demonstrated relatively good performance in the investigated scenario.
Its accuracy when tasked with counting people wearing protective helmets was found to be sufficient, and was validated against a different system.
A number of discrepancies between the counts of the model and the camera can be attributed to unexpected situations (Figs.~\ref{fig:worker1} and~\ref{fig:worker7}) and the fact that the Dahua camera did not differentiate people wearing and not wearing helmets.
The high correlation coefficient between the camera and the Orange AI\&ML Platform's model allows to conclude that the two solutions perform comparably well.

It should be noted that there were changes in the correlation between the days of experiments.
These differences are explained by the changes to the bounding box. 
This is one of the parameters that have to be investigated further.

Both variants of the proposed architecture can be used in the investigated scenario of PPE detection on a construction site.
The feasibility of using an edge-deployment was confirmed -- the server's computational capabilities were sufficient to maintain satisfactory inference accuracy.
Therefore, it can be concluded that the construction site is equipped with sufficient hardware to warrant further experiments with the deployment.

In the future, the two proposed architecture variants will be compared in terms of network latencies, resource utilization, and their accuracy.
The presented model will also be tested further, which will include manually annotating the videos to obtain a ground truth for comparison.
This will allow for determining the actual accuracy of the developed model.
Further optimization of bounding box locations is also planned.

\begin{acks}
Work supported by ASSIST-IoT project funded from the European Union's H2020 RIA program under grant 957258.
\end{acks}

\bibliographystyle{ACM-Reference-Format}
\bibliography{references}
\end{document}